\pdfoutput=1
\documentclass[11pt]{article}

\usepackage{ACL2023}

\usepackage{times}
\usepackage{latexsym}

\usepackage[T1]{fontenc}
\usepackage{graphicx}
\usepackage{float}
\usepackage{color, colortbl}
\usepackage{xcolor}
\definecolor{Gray}{gray}{0.9}
\usepackage{makecell}
\usepackage{lipsum}
\usepackage[most]{tcolorbox}
\usepackage{float}
\usepackage{verbatim} %In the preamble 
\usepackage{placeins}
\usepackage{makecell}
\usepackage{epstopdf}
\usepackage{algorithm}
\usepackage{amsmath}
\usepackage{amssymb}
\usepackage{pifont}
\usepackage{multirow}
\usepackage{subcaption}

\usepackage[utf8]{inputenc}

\usepackage{microtype}

\usepackage{inconsolata}

\usepackage{hyperref}
\usepackage{xurl}

\title{COSMMIC: Comment-Sensitive Multimodal Multilingual Indian Corpus for Summarization and Headline Generation}
\author{
  \textbf{Raghvendra Kumar\textsuperscript{1}},
  \textbf{S. A. Mohammed Salman\textsuperscript{2}},
  \textbf{Aryan Sahu\textsuperscript{3}},
  \textbf{Tridib Nandi\textsuperscript{4}}, \\
  \textbf{Pragathi Y. P.\textsuperscript{5}},
  \textbf{Sriparna Saha\textsuperscript{1}},
  \textbf{Jose G. Moreno\textsuperscript{6}}
  \\
  \\
  \textsuperscript{1}Department of Computer Science and Engineering, Indian Institute of Technology Patna, India \\
  \textsuperscript{2}Department of Metallurgical and Materials Engineering, National Institute of Technology Tiruchirappalli, India \\
  \textsuperscript{3}Department of Computer Science and Information Systems, BITS Pilani – Goa Campus, India \\
  \textsuperscript{4}Department of Computer Science and Engineering, Indian Institute of Information Technology Vadodara, India \\
  \textsuperscript{5}Department of Computer Science and Engineering, B.M.S. College of Engineering, Bangalore, India \\
  \textsuperscript{6}Université de Toulouse, IRIT UMR 5505 CNRS, France
  \\
  \\
  \small{
  \textbf{For inquiries:} 
  \{\texttt{raghvendra.kumar1004, samohammedsalman, aryansahu010103\}@gmail.com}}\\
  \small{
  \textbf{For inquiries:}
  \{\texttt{tridib.nandi25062002, pragathiyp2352, sriparna.saha\}@gmail.com}}\\
  \small{
  \textbf{For inquiries:}
  \texttt{jose.moreno@irit.fr}
  }
}

\begin{document}

\maketitle
\begin{abstract}
Despite progress in comment-aware multimodal and multilingual summarization for English and Chinese, research in Indian languages remains limited. This study addresses this gap by introducing COSMMIC, a pioneering comment-sensitive multimodal, multilingual dataset featuring nine major Indian languages. COSMMIC comprises 4,959 article-image pairs and 24,484 reader comments, with ground-truth summaries available in all included languages. Our approach enhances summaries by integrating reader insights and feedback. We explore summarization and headline generation across four configurations: (1) using article text alone, (2) incorporating user comments, (3) utilizing images, and (4) combining text, comments, and images. To assess the dataset’s effectiveness, we employ state-of-the-art language models such as LLama3 and GPT-4. We conduct a comprehensive study to evaluate different component combinations, including identifying supportive comments, filtering out noise using a dedicated comment classifier using IndicBERT, and extracting valuable insights from images with a multilingual CLIP-based classifier. This helps determine the most effective configurations for natural language generation (NLG) tasks. Unlike many existing datasets that are either text-only or lack user comments in multimodal settings, COSMMIC uniquely integrates text, images, and user feedback. This holistic approach bridges gaps in Indian language resources, advancing NLP research and fostering inclusivity. The resources are available at: \url{https://github.com/AaryanSahu/COSMMIC}.
\end{abstract}

\section{Introduction}
In today’s digital era, online content seamlessly integrates text and visuals, with user comments enriching discussions and providing diverse perspectives. \textbf{Reader-Aware Comment-Based Summarization (RACBS)} harnesses these interactions to generate more relevant and comprehensive summaries by incorporating user feedback \citep{10.1145/3583780.3614849}. However, the development of multilingual, comment-based, multimodal datasets for Indian languages remains largely unexplored. For instance, a news article on the Indian general elections\footnote{\url{https://en.wikipedia.org/wiki/2024_Indian_general_election}} often features images alongside a flood of reader comments in multiple languages. While the text delivers factual information and images offer visual context, the comments reflect public sentiment, debates, and interpretations, making them a valuable yet underutilized resource for enhancing summarization.

\textbf{Motivation for the dataset:} The challenge lies in analyzing such multifaceted content comprehensively. Current datasets often fail to integrate all these elements—text, images, and comments—particularly in the context of Indian languages. A dataset limited to text or images alone cannot capture the full spectrum of user engagement and content richness. Our dataset COSMMIC addresses this shortcoming by combining article text, images, and user comments in \textbf{nine major Indian languages: \textit{Bengali, Hindi, Gujarati, Marathi, Malayalam, Odia, Tamil, Telugu, and Kannada.}} This integrated approach offers a more complete view of online content. It is essential for developing advanced NLP models that can process and understand the full scope of information in today’s digital media landscape.

\newcommand{\Checkmark}{\ding{51}}  % Tick symbol
\newcommand{\XSolidBrush}{\ding{55}}  % Cross symbol

\begin{table*}[t]
    \centering
    \resizebox{\textwidth}{!}{
        \begin{tabular}{lcccccccccc}
        \hline
            \textbf{Datasets:} & \textbf{MMSUM} & \textbf{M2DS} & \textbf{Varta} & \textbf{XLHT} & \textbf{INLG} & \textbf{MSMO} & \textbf{M3LS} & \textbf{RAMDS} & \textbf{RASG} & \textbf{COSMMIC} \\ \hline
            \textbf{Multilingual} & \Checkmark & \Checkmark & \Checkmark & \Checkmark & \Checkmark & \XSolidBrush & \Checkmark & \XSolidBrush & \XSolidBrush & \Checkmark \\ 
            \textbf{Multimodal} & \Checkmark & \XSolidBrush & \XSolidBrush & \Checkmark & \XSolidBrush & \Checkmark & \Checkmark & \XSolidBrush & \XSolidBrush & \Checkmark \\ 
            \textbf{Comments} & \XSolidBrush & \XSolidBrush & \XSolidBrush & \XSolidBrush & \XSolidBrush & \XSolidBrush & \XSolidBrush & \Checkmark & \Checkmark & \Checkmark \\ 
            \textbf{Summaries} & \Checkmark & \Checkmark & \XSolidBrush & \XSolidBrush & \Checkmark & \Checkmark & \Checkmark & \Checkmark & \Checkmark & \Checkmark \\ 
            \textbf{Headlines} & \XSolidBrush & \XSolidBrush & \Checkmark & \Checkmark & \Checkmark & \Checkmark & \Checkmark & \Checkmark & \XSolidBrush & \Checkmark \\
            \textbf{\# Indian Lang.} & \textbf{8} & \textbf{1} & \textbf{14} & \textbf{8} & \textbf{11} & \textbf{0} & \textbf{8} & \textbf{0} & \textbf{0} & \textbf{9} \\ \hline
        \end{tabular}
    }
    \caption{Comparison of Datasets Based on Multilingualism, Multimodality, Comment Inclusion, Presence of Summaries and Headlines, and Coverage of Indian Languages.}
    \label{tab:comparison}
\end{table*}

\textbf{NLG tasks driven by the dataset presented:} Summarization and headline generation are key NLP challenges with applications in news aggregation, content curation, and information retrieval \citep{tan2017neural,singh2021sheg,verma2023synthesized,hamborg2017matrix,khatter2022content,gaci2023targeting,mahalakshmi2022summarization,esteva2021covid}. Summarization condenses text into concise, coherent outputs, while headline generation crafts engaging titles, both leveraging advanced language models \citep{zhang2024benchmarking} and multimodal data \citep{krubinski2024towards} for improved accuracy and relevance.  

Integrating images and user comments enriches these tasks, adding context to summaries and enhancing headline relevance \citep{li2017reader,gao2019abstractive}. Multimodal approaches \citep{he2023align,qiao2022grafting} combining text, visuals, and user interactions produce more informative outputs. Building on these advancements, we assess the impact of filtered comments and categorized images on NLG tasks. By integrating only informative inputs, our approach enhances coherence and relevance, with evaluations confirming its effectiveness in multilingual and multimodal RACBS. Our \textbf{contributions} enhance context-aware summaries and headlines, as detailed below.

\emph{\textbf{(1)} We introduce ``COSMMIC", an innovative comment-sensitive multimodal, multilingual dataset having article-image pairs with reader comments covering nine major Indian languages: Bengali, Hindi, Gujarati, Marathi, Malayalam, Odia, Tamil, Telugu, and Kannada. This dataset represents a pioneering effort in this field.}

\emph{\textbf{(2)} Human-written ground truth summaries are provided for all multilingual news articles within the dataset.}

\emph{\textbf{(3)} We establish benchmarks for our dataset by evaluating two NLG tasks—headline generation and summarization—using state-of-the-art language models such as GPT-4 and LLama3.}

\emph{\textbf{(4)} An extensive study is conducted to explore various configurations: article text alone, article text with comments, article text with relevant comments, article text with images, article text with images and comments, and article text with images and relevant comments.}

\emph{\textbf{(5)} To advance this task, we propose methods for filtering noisy comments and classifying images in the Hindi dataset. Using IndicBERT \citep{kakwani2020indicnlpsuite}, we develop a classifier to retain only informative comments and filter out noisy comments. Additionally, a threshold-based multilingual CLIP \citep{carlsson2022cross,reimers-2019-sentence-bert} classifier categorizes images as \emph{supplementary} (reinforcing the text) or \emph{complementary} (providing additional context). To assess linguistic preservation in RACBS, we perform a back-translation study on the Hindi dataset, highlighting quality degradation and the need to retain the original language.}

\section{Related Works}
We explore multilingual, multimodal datasets with reader comments for NLG tasks, as Table \ref{tab:comparison} details their linguistic, modal, and structural differences, excluding English from Indian languages.

\textbf{Multilingual Datasets:} 
\citet{scialom2020mlsum} introduced MLSUM, a large-scale dataset containing 1.5 million article-summary pairs in five languages, emphasizing cross-lingual biases. \citet{hasan2021xl} developed XL-Sum, featuring 1 million pairs across 44 languages for abstractive summarization. \citet{ladhak2020wikilingua} presented WikiLingua, which includes 18 languages and image-based alignments for cross-lingual summarization. \citet{hewapathirana2024m2ds} introduced M2DS, the first multilingual dataset for multi-document summarization (MDS) using BBC articles. \citet{aralikatte2023varta} developed Vārta, a dataset designed for headline generation in 14 Indic languages and English. \citet{kumar-etal-2022-indicnlg} introduced the IndicNLG (abbreviated as INLG in Table \ref{tab:comparison}) benchmark for NLG tasks across 11 Indic languages. Readers are also referred to notable works such as \citet{urlana-etal-2023-pmindiasum}, \citet{cao2020multisumm}, \citet{varab2021massivesumm}, \citet{feng2022msamsum}, and \citet{gliwa2019samsum}.

\textbf{Multimodal Datasets:} In the realm of multimodal datasets, \citet{zhu-etal-2018-msmo} introduced MSMO, a dataset for multimodal summarization that includes both text and image outputs and features a multimodal attention model. \citet{verma2023large} developed M3LS, the largest multilingual multimodal summarization dataset, comprising over a million document-image pairs across 20 languages. \citet{krubinski-pecina-2024-towards} presented a unified encoder-decoder model for headline generation, which enhances M3LS by incorporating additional images. \citet{wang-etal-2022-n24news} released N24News, a dataset from the New York Times with both text and image information organized into 24 categories. \citet{shohan2024xl} developed XL-HeadTags (shortened as XLHT in Table \ref{tab:comparison}), a multilingual and multimodal dataset for headline and tag generation, using images and captions in 20 languages. \citet{gotmare2024multimodal} proposed a machine-learning algorithm designed to generate news articles from geo-tagged images. \citet{liang2023summary}  introduced MM-Sum, a multilingual multimodal abstractive summarization dataset enhancing summary quality across 44 languages. Other noteworthy works include \citet{kumar2024multilingual}, \citet{mita2023camera}, \citet{fu-etal-2021-mm}, \citet{patil2024refinesumm}, \citet{qiu2024mmsum}, \citet{nguyen2023loralay}, and \citet{ghosh2024clipsyntel}. \emph{\textbf{Limitations:} Despite their significance, these datasets are text-only or multimodal without user comments.}

\textbf{Reader Comments Inclusive Datasets:} 
\citet{zaidan-callison-burch-2011-arabic} pioneered the Topic-driven Reader Comments Summarization (Torcs) problem by creating a dataset of 1,005 Yahoo! News articles with over one million comments. \citet{li2017reader} extended a variational autoencoder-based framework for reader-aware multi-document summarization (RA-MDS), presenting a new dataset with a detailed collection, annotation, and expert review process, demonstrating improved summarization performance with reader comments. \citet{10.1145/3340531.3412764} introduced a dataset of journalist-reader dialogues from The Guardian, proposing a ranking task for comments. \citet{gao2019abstractive} developed the Reader-Aware Summary Generator (RASG) using reader comments from Weibo to enhance summary relevance. Other noteworthy works include \citet{zaidan-callison-burch-2011-arabic}, \citet{10.1145/3589334.3645677}, \citet{fujita-etal-2019-dataset}, \citet{roha2022unsupervised}, and \citet{8419211}. \emph{\textbf{Limitations:} Despite their significance, these datasets lack focus on Indian languages, highlighting the need for our proposed dataset.}

\section{COSMMIC Dataset}
Our dataset is pioneering in the Indian context, featuring 4,959 article-image pairs and 24,484 reader comments, with comprehensive ground-truth summaries across all languages. It encompasses 500 to 650 articles for each of the nine major languages: \emph{Bengali, Hindi, Gujarati, Marathi, Malayalam, Odia, Tamil, Telugu, and Kannada.} While the dataset might seem relatively small for each individual language, it offers a significant volume compared to the RAMDS benchmark dataset \cite{li2017reader}, which includes 450 news articles and 10,000 comments across 45 topics. Our dataset provides broader coverage across nine languages, enhancing its utility. It also enabled us to manually verify each article to ensure that, after scraping and processing, the dataset is properly structured and free from any irrelevant information.

Additionally, with the rise of LLMs and their emphasis on zero-shot \citep{kojima2022large} and few-shot learning \citep{brown2020language}, the size of our dataset is well-suited for fine-tuning these models. \emph{Moreover, our dataset encompasses news exclusively from 2024. Extending further back proved impractical due to the \textbf{limited number of comments on older articles.} Even for articles from 2023 and 2022, there were only a few comments per 100 articles. Consequently, we concentrated on more recent news to ensure a robust dataset.}

\subsection{Dataset Curation}
\textbf{Data Source:} We selected DailyHunt for data scraping due to its diverse, multilingual content and integration of text, images, and user comments, providing a rich multimodal dataset for tasks like summarization and headline generation. Its focus on regional languages and local news fills gaps in predominantly English-centric datasets, while its real-time updates enable applications such as trend analysis and fake news detection. Additionally, the platform's vast articles and user interactions enable scalable research and model training, making it invaluable to our study.

\begin{figure}[t]
\centering
\includegraphics[width=1\columnwidth]{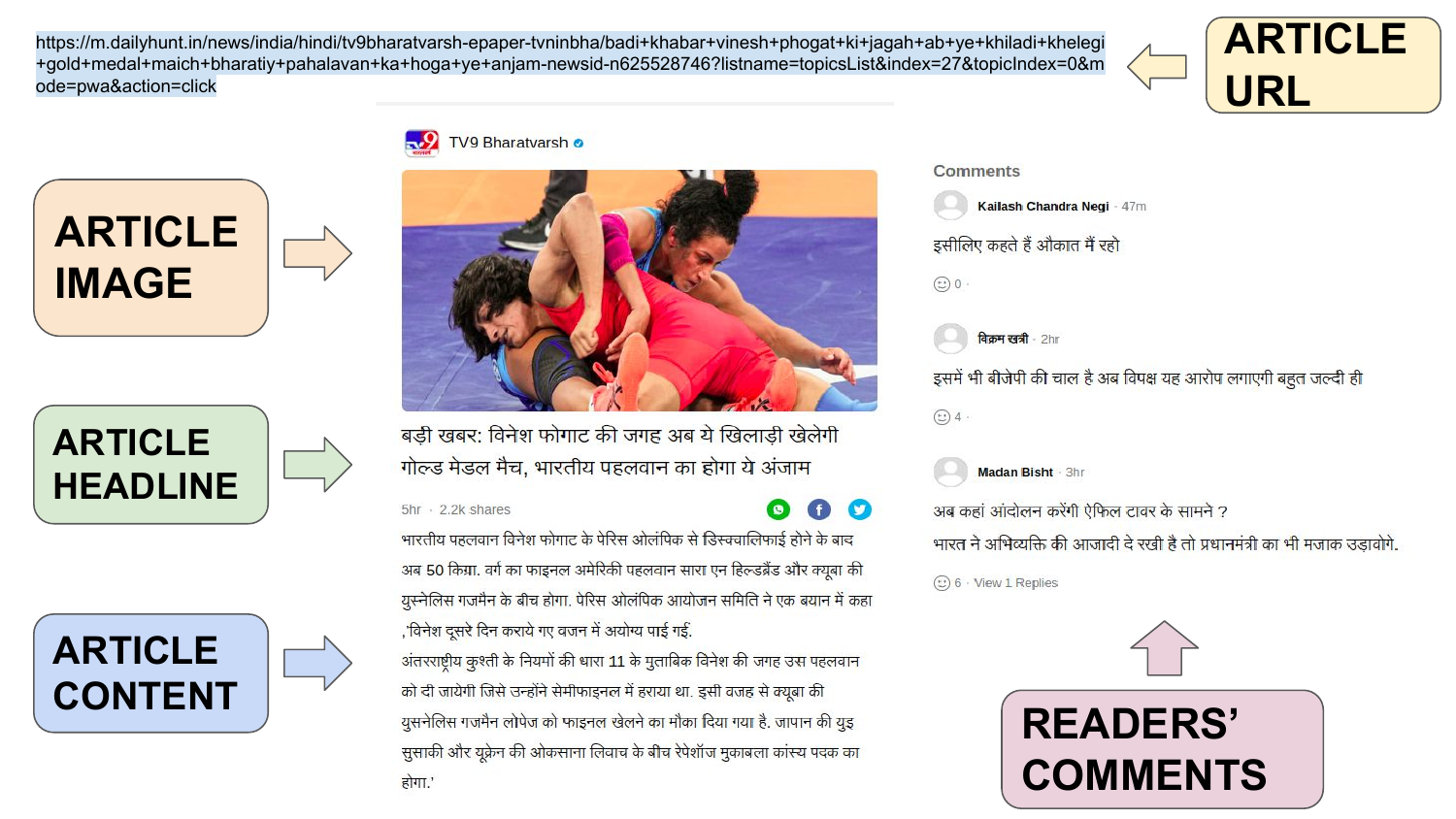} 
\caption{Various components extracted from a news article on DailyHunt.}
\label{fig:website}
\end{figure}

\textbf{Scrapping tools:} We used Selenium to scrape news articles from DailyHunt by navigating URLs, scrolling to load dynamic content, and collecting unique ``Read more" links. For each article with comments, we extracted the headline, related images, main content (excluding disclaimers), and comments, saving the data to a CSV file. The process involved initializing the Chrome WebDriver, dynamically loading content, and using BeautifulSoup for HTML parsing. \emph{Figure \ref{fig:website} illustrates the extracted components for a sample article.} The language-agnostic code was uniformly applied across nine languages. After scraping, we manually verified and cleaned the data, removing author and editor names (1-2\% of cases) for consistency.

\subsection{Statistics}
We present a quantitative overview of our dataset. \emph{Table \ref{tab:stats}} details key attributes per article, \emph{Table \ref{tab:stats-count}} lists article and comment counts by language, and \emph{Table \ref{tab:head-sum}} compares average word and sentence counts for headlines and summaries. Marathi has the highest comment volume and density, while Odia has the lowest. Bengali and Marathi lead in article counts, with Marathi, Bengali, and Hindi showing the most comment activity, reflecting higher user interaction.

\begin{table*}[!ht]
    \centering
    \scalebox{0.95}
    {
    \begin{tabular}{lrrrrrrrrr}
    \hline
        \textbf{Languages $\rightarrow$} & \multicolumn{3}{c}{\textbf{Hindi}} & \multicolumn{3}{c}{\textbf{Bengali}} & \multicolumn{3}{c}{\textbf{Marathi}} \\ \hline
        \textbf{Features $\downarrow$} & \textbf{\textit{Min}}& \textbf{\textit{Max}}& \textbf{\textit{Average}} & \textbf{\textit{Min}}& \textbf{\textit{Max}}& \textbf{\textit{Average}} & \textbf{\textit{Min}}& \textbf{\textit{Max}}& \textbf{\textit{Average}} \\ \hline
        \textbf{Token} & 118 & 4160 & 518 & 28 & 1962 & 339 & 36 & 2652 & 353 \\
        \textbf{Words} & 100 & 3712 & 466 & 24 & 1661 & 288 & 31 & 2229 & 303 \\ 
        \textbf{Unique Words} & 69 & 966 & 214 & 23 & 834 & 193 & 24 & 1079 & 202 \\ 
        \textbf{Sentences} & 2 & 228 & 25 & 2 & 119 & 26 & 2 & 210 & 24 \\ 
        \textbf{Comments} & 1 & 148 & 6 & 1 & 107 & 6 & 1 & 153 & 6 \\ \hline
        \textbf{Languages $\rightarrow$} & \multicolumn{3}{c}{\textbf{Gujarati}} & \multicolumn{3}{c}{\textbf{Malayalam}} & \multicolumn{3}{c}{\textbf{Odia}} \\ \hline
        \textbf{Features $\downarrow$} & \textbf{\textit{Min}}& \textbf{\textit{Max}}& \textbf{\textit{Average}} & \textbf{\textit{Min}}& \textbf{\textit{Max}}& \textbf{\textit{Average}} & \textbf{\textit{Min}}& \textbf{\textit{Max}}& \textbf{\textit{Average}} \\ \hline
        \textbf{Token} & 45 & 2254 & 410 & 31 & 2005 & 259 & 46 & 3058 & 281 \\
        \textbf{Words} & 43 & 2027 & 362 & 27 & 1666 & 218 & 41 & 2511 & 242 \\
        \textbf{Unique Words} & 40 & 1010 & 212 & 24 & 1186 & 172 & 34 & 1156 & 161 \\ 
        \textbf{Sentences} & 1 & 137 & 23 & 1 & 186 & 23 & 2 & 297 & 20 \\ 
        \textbf{Comments} & 1 & 34 & 3 & 1 & 82 & 6 & 1 & 25 & 2 \\ \hline
        \textbf{Languages $\rightarrow$} & \multicolumn{3}{c}{\textbf{Tamil}} & \multicolumn{3}{c}{\textbf{Telugu}} & \multicolumn{3}{c}{\textbf{Kannada}} \\ \hline
        \textbf{Features $\downarrow$} & \textbf{\textit{Min}}& \textbf{\textit{Max}}& \textbf{\textit{Average}} & \textbf{\textit{Min}}& \textbf{\textit{Max}}& \textbf{\textit{Average}} & \textbf{\textit{Min}}& \textbf{\textit{Max}}& \textbf{\textit{Average}} \\ \hline
        \textbf{Token} & 37 & 1882 & 289 & 48 & 1918 & 327 & 53 & 1641 & 258 \\
        \textbf{Words} & 28 & 1403 & 239 & 40 & 1651 & 268 & 43 & 1402 & 215 \\
        \textbf{Unique Words} & 24 & 866 & 171 & 38 & 1113 & 191 & 39 & 820 & 155 \\
        \textbf{Sentences} & 2 & 137 & 20 & 1 & 245 & 28 & 2 & 189 & 19 \\
        \textbf{Comments} & 1 & 93 & 6 & 1 & 103 & 4 & 1 & 62 & 5 \\ \hline
    \end{tabular}
    }
    \caption{Statistical Summary of Feature Counts across Various Indian Languages in the COSMMIC Dataset. Note that the average values have been rounded to the nearest integer.}
    \label{tab:stats}
\end{table*}

\begin{table}[!ht]
    \centering
    \scalebox{0.95}
    {
    \begin{tabular}{lrr}
    \hline
        \textbf{Language} & \textbf{\# Articles} & \textbf{\# Comments} \\ \hline
        \textbf{Hindi} & 543 & 3372 \\ 
        \textbf{Bengali} & 627 & 3742 \\ 
        \textbf{Marathi} & 650 & 3870 \\ 
        \textbf{Gujarati} & 537 & 1790 \\ 
        \textbf{Malayalam} & 505 & 2833 \\ 
        \textbf{Odia} & 577 & 1241 \\ 
        \textbf{Tamil} & 505 & 3204 \\ 
        \textbf{Telugu} & 505 & 1774 \\ 
        \textbf{Kannada} & 510 & 2658 \\ \hline
        \textbf{Total} & 4959 & 24484 \\ \hline
    \end{tabular}
    }
    \caption{Distribution of Articles and Comments Across Indian Languages in the COSMMIC Dataset.}
    \label{tab:stats-count}
\end{table}

\begin{table}[!ht]
    \centering
    \scalebox{0.85}
    {
    \begin{tabular}{lrrrr}
    \hline
        \textbf{Attribute} & \multicolumn{2}{c}{\textbf{Headlines}} & \multicolumn{2}{c}{\textbf{Summaries}} \\ \hline
        \textbf{Language} & \textbf{\textit{\# Words}} & \textbf{\textit{\# Sent.}} & \textbf{\textit{\# Words}} & \textbf{\textit{\# Sent.}} \\ \hline
        \textbf{Hindi} & 18 & 1 & 117 & 6 \\ 
        \textbf{Bengali} & 12 & 1 & 86 & 6 \\ 
        \textbf{Marathi} & 13 & 1 & 110 & 7 \\ 
        \textbf{Gujarati} & 13 & 1 & 90 & 6 \\ 
        \textbf{Malayalam} & 12 & 1 & 59 & 6 \\ 
        \textbf{Odia} & 10 & 1 & 83 & 6 \\ 
        \textbf{Tamil} & 11 & 2 & 77 & 6 \\ 
        \textbf{Telugu} & 8 & 1 & 90 & 9 \\
        \textbf{Kannada} & 10 & 1 & 66 & 5 \\ \hline
    \end{tabular}
    }
    \caption{Average Word and Sentence Counts for Headlines and Summaries (rounded to the nearest integer).}
    \label{tab:head-sum}
\end{table}

\subsection{Crafting Summaries}
\textbf{Flexibility in Summary Composition:} Substantial variability in article features across different Indian languages is observed, as detailed in Table \ref{tab:stats}, underscoring the necessity of flexibility in summary composition. Imposing a rigid summary length, such as a fixed word count \cite{li2017reader}, would not have accommodated the natural linguistic diversity and structural differences among these languages. A one-size-fits-all approach would have risked compromising the quality and relevance of the summaries, failing to capture the true essence of the content. Therefore, flexibility in summary length and structure was crucial to accurately capture the unique characteristics and information density of each language, a principle central to our approach. \textbf{The summaries created were inclusive of user comments.} Noisy comments were manually removed (\emph{in appendix~\ref{mfn}}).

\textbf{Annotator Guidelines:} A diligently selected team of ten annotators, each holding a master's in linguistics for their respective languages, was engaged to ensure the highest standards in creating ground truth summaries. Each annotator was proficient in at least three languages, allowing for a profound understanding of linguistic sharpness and cultural contexts across the nine languages in the dataset. The process involved four annotators per language: two generated the summaries, with each summarizing half of the sample set, such as 50 out of 100 samples, while the remaining two conducted thorough reviews of all summaries. A summary was deemed final only when it received approval from both reviewing annotators, ensuring a robust quality assurance process. The guidelines mandated that summaries be concise yet comprehensive, capturing essential ideas and significant details while accommodating natural variability in length. Annotators were instructed to faithfully reflect the meaning and context of the original text, avoiding biases and preserving the tone and intent of the source material. Importantly, reader comments were excluded from consideration to prevent external influences. Adhering to these standards and employing expert annotators ensured the summaries were of high quality and preserved the dataset's integrity. We compensated annotators at USD 1.20 per 10 summaries.

\section{Summarization and Headline Generation} 

\subsection{Methodology} 

While prior studies \citet{ahuja2023mega,ahuja2024megaverse,kumar-etal-2022-indicnlg,singh2024indicgenbench,bhat-etal-2023-generative} have benchmarked NLG for Indic languages using LLMs, none have explored reader-aware summarization or headline generation, a gap our work addresses.

\textbf{Configurations Used:} Our comprehensive studies explore various configurations to assess the COSMMIC dataset for headline generation and summarization, utilizing both GPT-4 and LLAMA-3 models. The configurations included (1) using only the article text, (2) combining the article text with comments, (3) integrating the article text with relevant comments, (4) incorporating images along with the article text, (5) merging article text with both images and comments, and (6) combining article text, images, and relevant comments. These configurations were rigorously tested to assess how different data combinations influence the models' performance in generating accurate and contextually rich headlines and summaries across the various languages.

\textbf{Models Used:} We employed GPT-4 \citep{achiam2023gpt} and LLAMA-3 \citep{dubey2024llama}, state-of-the-art NLP models. GPT-4, a paid multimodal model, excels in processing text and images, making it ideal for our dataset. LLAMA-3, an open-source alternative with 8 billion parameters, offers strong language generation capabilities. Recent studies \citet{katz2024gpt,waisberg2023gpt} highlight GPT-4’s superiority in complex tasks, while our evaluation of LLAMA-3 provides insights into the performance of a lighter, cost-effective model for reader-aware summarization and headline generation. 

\textbf{Evaluation Metrics:} We assessed NLG tasks using ROUGE \citep{lin-2004-rouge} for n-gram overlap and BERTScore \citep{zhang2019bertscore} for semantic similarity, ensuring precise n-gram segmentation in Indic languages with the Polyglot tokenizer \citep{al-rfou-etal-2013-polyglot}. \emph{While these metrics may appear limited, they are the standard for state-of-the-art datasets (Table \ref{tab:comparison}) and provide a rigorous benchmark. Their widespread adoption reinforces their reliability, making them well-suited for our dataset evaluation.} 

% The full dataset will be publicly released, with a subset included in the supplementary materials.

\begin{table*}[!ht]
    \centering
    \scalebox{0.90}
    {
    \begin{tabular}{lrrrrrrrrrrrr}
    \hline
        \multicolumn{13}{c}{\textbf{Reader Aware Summarization with GPT-4}}  \\ \hline
        \textbf{} & \multicolumn{4}{c}{\textbf{Only Article}} & \multicolumn{4}{c}{\textbf{Article + Comments}} & \multicolumn{4}{c}{\textbf{Article + Rel. Com.}}  \\ \hline
        \textbf{Lang} & R1 & R2 & RL & BS & R1 & R2 & RL & BS & R1 & R2 & RL & BS \\ \hline
        \textbf{Hin.} & 44.18 & 18.94 & 33.47 & 76.61 & 50.98 & 24.82 & 38.34 & 78.22 & 52.13 & 25.52 & 39.43 & 78.71 \\ 
        \textbf{Ben.} & 29.40 & 9.32 & 21.52 & 74.69 & 36.64 & 13.38 & 26.15 & 76.94 & 37.19 & 13.73 & 26.82 & 77.34 \\ 
        \textbf{Mar.} & 27.24 & 10.49 & 25.79 & 74.28 & 40.38 & 18.56 & 37.97 & 78.02 & 41.50 & 19.47 & 39.03 & 78.56 \\ 
        \textbf{Guj.} & 26.97 & 8.38 & 25.13 & 75.00 & 32.00 & 12.00 & 29.46 & 76.26 & 32.89 & 12.41 & 30.29 & 76.63 \\ 
        \textbf{Mal.} & 17.38 & 5.45 & 16.55 & 72.78 & 18.88 & 6.16 & 17.90 & 73.49 & 20.18 & \textbf{6.71} & 19.11 & 74.01 \\ 
        \textbf{Odia.} & 25.45 & 7.98 & 18.90 & 92.47 & 34.99 & 13.73 & 26.47 & 92.91 & 36.36 & 14.11 & 27.68 & 92.83 \\ 
        \textbf{Tam.} & 25.07 & 9.52 & 23.60 & 74.05 & 32.63 & 14.32 & 30.67 & 77.20 & 34.94 & 15.52 & 32.73 & 77.83 \\ 
        \textbf{Tel.} & 21.91 & 6.83 & 20.79 & 73.05 & 34.32 & 13.89 & 32.35 & 76.80 & \textbf{36.19} & \textbf{15.07} & \textbf{33.93} & \textbf{77.75} \\ 
        \textbf{Kan.} & 23.94 & 8.04 & 22.70 & 74.83 & 28.85 & 10.68 & 27.08 & 75.89 & 29.10 & 10.79 & 28.01 & 76.01 \\ \hline
        \textbf{} & \multicolumn{4}{c}{\textbf{Article + Images}} & \multicolumn{4}{c}{\textbf{Article + Images + Comments}} & \multicolumn{4}{c}{\textbf{Article + Images + Rel. Com.}} \\ \hline
        \textbf{Lang} & R1 & R2 & RL & BS & R1 & R2 & RL & BS & R1 & R2 & RL & BS \\ \hline
        \textbf{Hin.} & \textbf{53.41} & \textbf{26.10} & \textbf{40.41} & \textbf{79.38} & 49.63 & 23.67 & 36.66 & 77.69 & 50.90 & 24.42 & 37.64 & 78.15 \\ 
        \textbf{Ben.} & \textbf{38.16} & \textbf{14.16} & \textbf{27.62} & \textbf{77.87} & 34.90 & 12.53 & 24.34 & 76.49 & 36.12 & 13.20 & 25.63 & 76.89 \\ 
        \textbf{Mar.} & \textbf{42.83} & \textbf{20.11} & \textbf{40.38} & \textbf{79.03} & 39.50 & 18.02 & 37.43 & 77.85 & 40.62 & 18.57 & 38.27 & 78.28 \\ 
        \textbf{Guj.} & \textbf{33.74} & \textbf{12.60} & \textbf{31.15} & \textbf{77.20} & 31.06 & 11.15 & 28.66 & 76.17 & 31.78 & 11.52 & 29.38 & 76.30 \\ 
        \textbf{Mal.} & \textbf{20.27} & 6.53 & \textbf{19.22} & \textbf{74.15} & 18.47 & 5.73 & 17.60 & 73.64 &  19.00 & 5.90 & 18.04 &	73.92 \\ 
        \textbf{Odia.} & \textbf{36.31} & \textbf{14.22} & \textbf{27.81} & \textbf{93.40} & 33.06 & 12.36 & 24.59 & 91.86 & 34.60 & 13.21 & 25.82 & 92.29 \\ 
        \textbf{Tam.} & \textbf{35.87} & \textbf{15.94} & \textbf{33.78} & \textbf{78.32} & 31.85 & 13.34 & 29.89 & 76.83 & 32.96 & 14.05 & 31.01 & 77.28 \\ 
        \textbf{Tel.} & 35.33 & 14.03 & 33.17 & 77.54 & 32.61 & 12.51 & 30.61 & 76.52 &  33.45 & 13.15 & 31.56 & 76.82 \\ 
        \textbf{Kan.} & \textbf{30.15} & \textbf{11.12} & \textbf{28.28} & \textbf{76.55} & 27.81 & 9.89 & 26.28 & 75.43 & 28.60 & 10.50 & 26.88 & 75.68 \\ \hline
        \multicolumn{13}{c}{\textbf{Reader Aware Headline Generation with GPT-4}} \\ \hline
        \textbf{} & \multicolumn{4}{c}{\textbf{Only Article}} & \multicolumn{4}{c}{\textbf{Article + Comments}} & \multicolumn{4}{c}{\textbf{Article + Rel. Com.}}  \\ \hline
        \textbf{Lang} & R1 & R2 & RL & BS & R1 & R2 & RL & BS & R1 & R2 & RL & BS \\ \hline
        \textbf{Hin.} & \textbf{32.23} & \textbf{9.58} & \textbf{24.54} & \textbf{72.97} & 30.27 & 8.99 & 23.61 & 72.89 & 30.49 & 9.12 & 23.94 & 72.87 \\ 
        \textbf{Ben.} & \textbf{22.07} & 4.96 & 17.27 & 72.63 & 21.01 & 4.80 & 17.14 & 72.66 & 21.27 & \textbf{5.14} & \textbf{17.28} & \textbf{72.72} \\ 
        \textbf{Mar.} & \textbf{21.18} & \textbf{6.51} & \textbf{18.33} & \textbf{72.83} & 20.19 & 5.46 & 17.47 & 72.63 & 20.01 & 5.51 & 17.48 & 72.63 \\ 
        \textbf{Guj.} & \textbf{20.47} & \textbf{5.37} & \textbf{17.49} & 73.67 & 19.81 & 4.86 & 17.13 & \textbf{73.89} & 20.10 & 5.26 & 17.42 & 73.84 \\ 
        \textbf{Mal.} & \textbf{20.41} & \textbf{5.53} & \textbf{17.48} & \textbf{72.19} & 18.99 & 4.87 & 16.38 & 71.75 & 19.19 & 4.89 & 16.09 & 71.87 \\ 
        \textbf{Odia.} & 18.23 & 4.80 & 15.55 & 90.03 & 19.09 & \textbf{4.83} & 16.14 & \textbf{91.35} & \textbf{19.38} & 4.74 & \textbf{16.32} & 91.27 \\ 
        \textbf{Tam.} & \textbf{24.63} & \textbf{6.66} & \textbf{22.13} & \textbf{72.59} & 22.03 & 6.45 & 19.77 & 72.37 & 22.71 & 6.62 & 20.45 & 72.55 \\ 
        \textbf{Tel.} & \textbf{18.82} & 4.51 & \textbf{16.33} & 70.61 & 17.97 & \textbf{4.56} & 15.90 & 70.63 & 18.11 & 4.55 & 16.10 & \textbf{70.86} \\ 
        \textbf{Kan.} & 20.88 & 5.04 & 17.97 & 73.22 & 22.12 & 6.32 & 19.16 & 74.07 & \textbf{22.87} & \textbf{6.54} & \textbf{19.78} & \textbf{74.12} \\ \hline
        \textbf{} & \multicolumn{4}{c}{\textbf{Article + Images}} & \multicolumn{4}{c}{\textbf{Article + Images + Comments}} & \multicolumn{4}{c}{\textbf{Article + Images + Rel. Com.}} \\ \hline
        \textbf{Lang} & R1 & R2 & RL & BS & R1 & R2 & RL & BS & R1 & R2 & RL & BS \\ \hline
        \textbf{Hin.} & 28.56 & 8.31 & 22.44 & 72.34 & 29.14 & 8.53 & 22.88 & 72.29 & 28.69 & 8.37 & 22.64 & 72.38 \\ 
        \textbf{Ben.} & 18.21 & 3.66 & 14.71 & 72.24 & 18.67 & 3.63 & 14.85 & 72.11 & 19.45 & 4.04 & 15.81 & 72.22 \\ 
        \textbf{Mar.} & 18.18 & 4.49 & 15.59 & 72.26 & 19.03 & 5.02 & 16.34 & 72.25 & 19.07 & 4.67 & 16.45 & 72.30 \\ 
        \textbf{Guj.} & 15.98 & 3.52 & 13.83 & 72.51 & 16.42 & 3.43 & 14.25 & 72.47 & 16.86 & 3.56 & 14.51 & 72.65 \\ 
        \textbf{Mal.} & 14.80 & 2.78 & 12.92 & 70.40 & 15.18 & 3.45 & 13.02 & 70.37 & 15.41 & 3.53 & 13.25 & 70.43 \\ 
        \textbf{Odia.} & 15.89 & 3.43 & 13.32 & 90.11 & 16.76 & 3.84 & 14.18 & 90.23 & 16.94 & 3.89 & 14.53 & 90.69 \\ 
        \textbf{Tam.} & 18.93 & 4.57 & 17.18 & 71.81 & 20.03 & 4.36 & 17.99 & 71.56 & 20.31 & 4.74 & 18.42 & 71.80 \\ 
        \textbf{Tel.} & 15.44 & 3.31 & 14.01 & 70.28 & 16.96 & 3.84 & 15.21 & 70.46 & 17.85 & 4.36 & 16.15 & 70.64 \\ 
        \textbf{Kan.} & 19.46 & 4.35 & 16.95 & 72.92 & 19.27 & 4.63 & 17.04 & 73.08 & 19.62 & 5.05 & 17.37 & 73.28 \\ \hline
        
    \end{tabular}
    }
    \caption{Performance of Summarization and Headline Generation using GPT-4 across different languages and input combinations. The table shows ROUGE-1 (R1), ROUGE-2 (R2), ROUGE-L (RL), and BERTScore (BS) metrics. Bold values indicate the highest performance metrics achieved for each language and input type, highlighting the most effective combination.}
    \label{tab:gpt4}
\end{table*}

\begin{table*}[!ht]
    \centering
    \scalebox{0.90}
    {
    \begin{tabular}{lrrrrrrrrrrrr}
    \hline
        \multicolumn{13}{c}{\textbf{Reader Aware Summarization with Llama-3}}  \\ \hline
        \textbf{} & \multicolumn{4}{c}{\textbf{Only Article}} & \multicolumn{4}{c}{\textbf{Article + Comments}} & \multicolumn{4}{c}{\textbf{Article + Rel. Com.}}  \\ \hline
        \textbf{Lang} & R1 & R2 & RL & BS & R1 & R2 & RL & BS & R1 & R2 & RL & BS \\ \hline
         \textbf{Hin.} & \textbf{36.49} & \textbf{14.88} & \textbf{34.97} & \textbf{84.45} & 31.27 & 11.95 & 29.89 & 83.63 & 34.35 & 12.76 & 33.13 & 84.26 \\ 
        \textbf{Ben.} & 2.97 & 1.04 & \textbf{2.91} & 82.10 & \textbf{2.98} & \textbf{1.14} & 2.89 & \textbf{82.12} & 2.64 & 0.76 & 2.64 & 82.08 \\ \
        \textbf{Mar.} & 19.96 & \textbf{8.36} & \textbf{19.41} & \textbf{85.29} & 18.13 & 4.19 & 11.09 & 79.38 & \textbf{20.05} & 8.01 & 19.39 & 85.21 \\ \
        \textbf{Guj.} & \textbf{29.71} & \textbf{11.54} & \textbf{28.70} & \textbf{93.48} & 29.44 & 11.53 & 28.45 & 93.69 & 27.89 & 9.82 & 26.92 & 93.46 \\
        \textbf{Mal.} & \textbf{22.04} & \textbf{7.60} & \textbf{21.65} & \textbf{93.60} & 20.60 & 6.66 & 20.10 & 93.38 & 20.76 & 7.36 & 20.28 & 93.50 \\ 
        \textbf{Odia.} & \textbf{1.96} & \textbf{0.96} & \textbf{1.96} & \textbf{92.18} & 2.03 & 0.70 & 2.03 & 91.77 & 0.12 & 0.03 & 0.12 & 91.50 \\ 		
        \textbf{Tam.} & 33.61 & 12.35 & 32.51 & 86.91 & 32.41 & 13.11 & 31.23 & 86.93 & \textbf{34.01} & \textbf{13.72} & \textbf{33.19} & \textbf{87.01} \\ 
        \textbf{Tel.} & \textbf{26.21} & \textbf{11.37} & \textbf{25.49} & 92.01 & 22.11 & 10.3 & 21.59 & 91.89 & 21.94 & 9.88 & 21.43 & \textbf{92.25} \\ 
        \textbf{Kan.} & \textbf{25.34} & \textbf{8.81} & \textbf{24.36} & 93.29 & 23.98 & 8.71 & 23.47 & \textbf{93.46} & 23.57 & 8.36 & 22.86 & 93.38 \\ \hline
        \multicolumn{13}{c}{\textbf{Reader Aware Headline Generation with Llama-3}} \\ \hline
        \textbf{} & \multicolumn{4}{c}{\textbf{Only Article}} & \multicolumn{4}{c}{\textbf{Article + Comments}} & \multicolumn{4}{c}{\textbf{Article + Rel. Com.}}  \\ \hline
        \textbf{Lang} & R1 & R2 & RL & BS & R1 & R2 & RL & BS & R1 & R2 & RL & BS \\ \hline
         \textbf{Hin.} & \textbf{7.76} & \textbf{0.86} & \textbf{7.59} & \textbf{76.60} & 5.29 & 0.42 & 5.19 & 76.35 & 4.20 & 0.33 & 4.08 & 76.11 \\ 
        \textbf{Ben.} & \textbf{0.59} & \textbf{0.23} & \textbf{0.59} & \textbf{73.51} & 0.58 & 0.16 & 0.58 & 72.87 & 0.25 & 0.07 & 0.25 & 72.50 \\ 
        \textbf{Mar.} & \textbf{4.25} & \textbf{0.41} & \textbf{4.15} & \textbf{76.03} & 2.97 & 0.15 & 2.89 & 75.47 & 2.57 & 0.23 & 2.57 & 75.03 \\ 
        \textbf{Guj.} & \textbf{4.57} & \textbf{1.08} & \textbf{4.51} & 82.94 & 4.56 & 0.91 & 4.50 & \textbf{82.95} & 4.09 & 0.59 & 4.09 & 82.49 \\ 
        \textbf{Mal.} & \textbf{3.56} & \textbf{0.43} & \textbf{3.56} & 81.89 & 2.79 & 0.20 & 2.79 & \textbf{82.34} & 2.62 & 0.06 & 2.52 & 76.31 \\ 
        \textbf{Odia.} & \textbf{0.76} & 0.23 & \textbf{0.76} & 82.63 & 0.65 & 0.03 & 0.65 & \textbf{82.91} & 0.62 & \textbf{0.24} & 0.62 & 81.93 \\ 
        \textbf{Tam.} & 5.34 & \textbf{0.99} & 5.26 & 75.85 & \textbf{6.36} & 0.63 & \textbf{6.36} & \textbf{75.88} & 4.24 & 0.34 & 4.14 & 75.71 \\ 
        \textbf{Tel.} & \textbf{3.94} & \textbf{0.25} & \textbf{3.84} & 75.40 & 3.73 & \textbf{0.25} & 3.63 & 75.84 & 2.62 & 0.06 & 2.52 & \textbf{76.31} \\ 
        \textbf{Kan.} & \textbf{4.83} & \textbf{0.76} & \textbf{4.83} & 79.37 & 3.36 & 0.47 & 3.36 & \textbf{79.96} & 0.13 & 0.04 & 0.13 & 79.20 \\ \hline
        
    \end{tabular}
    }
    \caption{Performance of Summarization and Headline Generation using LLAMA-3 across languages and input types, showing ROUGE-1 (R1), ROUGE-2 (R2), ROUGE-L (RL), and BERTScore (BS) metrics. Bold values highlight the best results.}
    \label{tab:llama3}
\end{table*}

\subsection{Experiments, Results and Analysis}

\textbf{Key Prompting Strategies}
To enhance headline generation and summarization, we employ structured zero-shot prompts tailored for different multimodal scenarios. For headline generation, prompts ensure factual, concise, and engaging outputs, whether integrating images with article text or incorporating user comments as supplementary context. Similarly, summarization prompts maintain linguistic consistency while prioritizing core information, ensuring balanced and insightful summaries. \emph{Complete prompt structures and rationale are provided in the appendix~\ref{ps}.} 

\textbf{Analysis of Summarization with GPT-4} 

Table \ref{tab:gpt4} shows that ``Article + Images" generally provides the best summarization results across most Indian languages, improving both ROUGE and BERTScore metrics. Images offer valuable contextual support, especially in languages with complex scripts and structures, enhancing the understanding and interpretation of content. However, Telugu is an exception where ``Article + Images + Rel. Com." yields the highest scores, highlighting the importance of integrating relevant comments for improved summarization. This variation underscores the need for tailored summarization strategies that account for each language's unique linguistic and contextual features. \emph{The table further shows that incorporating \textbf{relevant comments enhances summarization quality}, as reflected in improved ROUGE and BERTScore metrics across most languages.} This reinforces the effectiveness of reader-aware summarization, where comments provide crucial context and sentiment, leading to more accurate summaries.

\textbf{Analysis of Headline Generation with GPT-4} While article-only inputs yield strong headlines, incorporating relevant comments enhances performance in certain languages by providing crucial context. \emph{Languages like Telugu and Kannada benefit significantly, demonstrating that \textbf{comments help bridge informational gaps and refine headlines.}} The minimal drop in BERTScore suggests that while images may not improve exact textual matches, they contribute valuable contextual insights. Overall, GPT-4 effectively optimizes NLG tasks across languages with diverse inputs.

\textbf{Summarization with LLama-3} 

Table \ref{tab:llama3} shows that ``Only Article" input generally yields the best summarization scores with LLAMA-3. However, incorporating relevant comments significantly boosts performance for languages like Tamil and Telugu, as seen in improved BERTScore metrics. For Bengali and Kannada, including all comments enhances contextual richness, leading to superior BERTScore results. This demonstrates that while minimal input is effective, relevant comments improve summary quality, even for lightweight models. 

\textbf{Headline Generation with LLama-3} 

While ``Only Article" inputs typically perform best for headline generation, Table \ref{tab:llama3} shows that adding all comments improves BERTScore across languages. This suggests that while simpler inputs work well, richer contextual data enhances semantic relevance and alignment. Thus, LLama-3 benefits from minimal inputs but achieves better headline quality with comments. \textbf{Human analysis is included in the appendix~\ref{ha} due to space constraints.}

\begin{table*}[htbp]
\centering
\scalebox{0.90}
{
\begin{tabular}{lrrrr|lrrrr}
\hline
\textbf{Config} & \textbf{R1} & \textbf{R2} & \textbf{RL} & \textbf{BS} & \textbf{Config} & \textbf{R1} & \textbf{R2} & \textbf{RL} & \textbf{BS} \\  
\hline
\textbf{OA} & 44.18 & 18.94 & 33.47 & 76.61 & \textbf{OA} & \textbf{32.23} & \textbf{9.58} & \textbf{24.54} & 72.97 \\  
\textbf{OA+SC} & 56.64 & 28.33 & 42.59 & 82.32 & \textbf{OA+SC} & 31.68 & 8.98 & 24.02 & 71.53 \\  
\textbf{OA+EC} & \textbf{58.22} & \textbf{29.19} & \textbf{43.93} & \textbf{84.15} & \textbf{OA+EC} & 32.06 & 9.52 & 23.89 & \textbf{73.44} \\  
\textbf{OA+DC} & 35.62 & 20.44 & 29.85 & 70.44 & \textbf{OA+DC} & 26.54 & 7.11 & 20.48 & 69.85 \\  
\textbf{OA+AC} & 50.98 & 24.82 & 38.34 & 78.22 & \textbf{OA+AC} & 30.27 & 8.99 & 23.61 & 72.89 \\  
\textbf{OA+RC (GPT)} & 52.13 & 25.52 & 39.43 & 78.71 & \textbf{OA+RC (GPT)} & 30.49 & 9.12 & 23.94 & 72.87 \\  
\hline
\end{tabular}
}
\caption{Performance of different configurations for \textbf{Summarization (left)} and \textbf{Headline Generation (right)} using GPT-4. Configurations: \textbf{OA} = Only Article, \textbf{SC} = Supporting Comments, \textbf{EC} = Enriching Comments, \textbf{DC} = Disconnected Comments, \textbf{AC} = All Comments, \textbf{RC (GPT)} = Relevant Comments selected by GPT.}
\label{tab:comments-classification-results}
\end{table*}

\section{Classifying Comments \& Images}
As an initial benchmark, we focus on the Hindi section of our dataset due to annotation costs and feasibility. Tasks below aim to enhance data quality and assess the impact of linguistic transformations. 

\textbf{Comment Classification:} To refine our dataset, we propose an automated approach to retain only informative comments while filtering out noisy ones in the Hindi section. Using IndicBERT \citep{kakwani2020indicnlpsuite}, we develop a classifier to enhance the quality of reader-aware summarization. For training, our annotator manually labelled 250 comments into three categories: \textbf{Supporting} (aligned with the article), \textbf{Enriching} (adding new insights), and \textbf{Disconnected} (irrelevant to the article). We fine-tuned IndicBERT on this labelled data and then applied the trained model to classify the remaining comments. Once categorized, we conducted further experiments to analyze how comment classification improves summarization performance, as shown in Table \ref{tab:comments-classification-results}. 

\textbf{Insight:} The results in Table~\ref{tab:comments-classification-results} indicate that integrating enriching comments (\textbf{OA+EC}) yields the highest improvement in summarization. Interestingly, supporting comments (\textbf{OA+SC}) also enhance performance but slightly less than enriching ones. Including disconnected comments (\textbf{OA+DC}) significantly deteriorates performance, highlighting the importance of filtering relevant contextual information. This trend continues for headline generation, where \textbf{OA+EC} outperforms other configurations, reinforcing the benefit of additional insights over mere agreement. \emph{A detailed discussion is furthermore provided in the appendix~\ref{cc} due to space constraints.}

\textbf{Image Classification:} We use a threshold-based classifier with multilingual CLIP \texttt{(sentence-transformers/ clip-ViT-B-32-multilingual-v1)} \citep{carlsson2022cross,reimers-2019-sentence-bert} to assess image relevance in multimodal summarization \citep{jangra2021multi}. Since images accompany articles from a reputable media house, classification into unrelated categories is unnecessary. Instead, we classify images as \emph{supplementary} (reinforcing the text) or \emph{complementary} (providing additional context). CLIPScore is computed using both the ground truth summary and the image, as well as the headline and the image. The average score determines classification: above 0.25 indicates reinforcement, while lower values suggest added context.

The 0.25 threshold is empirically set based on CLIPScore distributions in our dataset, ensuring balanced classification. Prior work \citet{jangra2021multi} shows real-world text-image pairs often have low CLIPScores, making higher thresholds (e.g., 0.5) impractical. This approach effectively differentiates between reinforcing and contextual images. We observed that 36.84\% of images reinforced the article (\emph{supplementary}), while most (63.16\%) added extra information (\emph{complementary}). This suggests that Hindi news articles are often accompanied by images that provide background information rather than directly supporting the text. As a result, multimodal summarization models must focus on interpreting contextual visuals rather than relying solely on direct text-image alignment (\textbf{further strengthening the findings of Table \ref{tab:gpt4}}). Likewise, headline generation may gain from complementary images for a broader context beyond the main points. 

\emph{Remark: For a deeper understanding, additional discussion and a detailed back-translation study focusing on preserving linguistic authenticity are provided in the appendix (see \ref{bp1} and \ref{bp2}). These sections explore the nuances of language preservation and offer insights into the effectiveness of our approach.}

\section{Conclusion}
In summary, the \textbf{COSMMIC} dataset is a notable advancement in multimodal, multilingual, and comment-sensitive resources for Indian languages. It integrates text, images, and reader comments across nine major languages, addressing gaps in existing datasets. This dataset enhances headline generation and summarization while offering insights into regional language dynamics and user interactions. Our ablation study demonstrates the effectiveness of different component combinations, making \textbf{COSMMIC} a valuable resource.

\section*{Limitations}  Although the dataset encompasses nine major Indian languages, it does not represent all regional dialects and languages within India. Also, the dataset's reliance on publicly available sources from DailyHunt means that the coverage may be skewed towards specific topics or viewpoints prevalent on the platform. 

\textbf{Ethics Statement} Our dataset, including URLs to original articles and images, content, ground truth summaries, and headlines, is publicly available on GitHub: \url{https://github.com/AaryanSahu/COSMMIC}, along with the scraping and experimental code. The scraping process deliberately excludes any user information to protect privacy.

\section*{Acknowledgments}
Raghvendra Kumar extends appreciation to the Prime Minister's Research Fellowship (PMRF) for its assistance and successful completion of this research. Sriparna Saha recognizes the support received from the ``Technology Innovation Hub, Vishlesan I-Hub Foundation IIT Patna" (Project No. TIH/CSE/ASMO/05) for the achievement of this research. Aryan Sahu sincerely thanks the India Chapter of ACM SIGKDD for the opportunity to participate in the Uplink Internship program, which played a pivotal role in the successful completion of this research work.

\bibliography{mybib}
\bibliographystyle{acl_natbib}

\appendix

\section{Appendix}

\subsection{Detailed Analysis of Comment Classification Impact}\label{cc}
The results in Table~\ref{tab:comments-classification-results} provide key insights into how different types of comments influence summarization and headline generation quality. Below, we analyze these findings in detail.

\begin{figure*}[htbp]
\centering
\includegraphics[width=\textwidth]{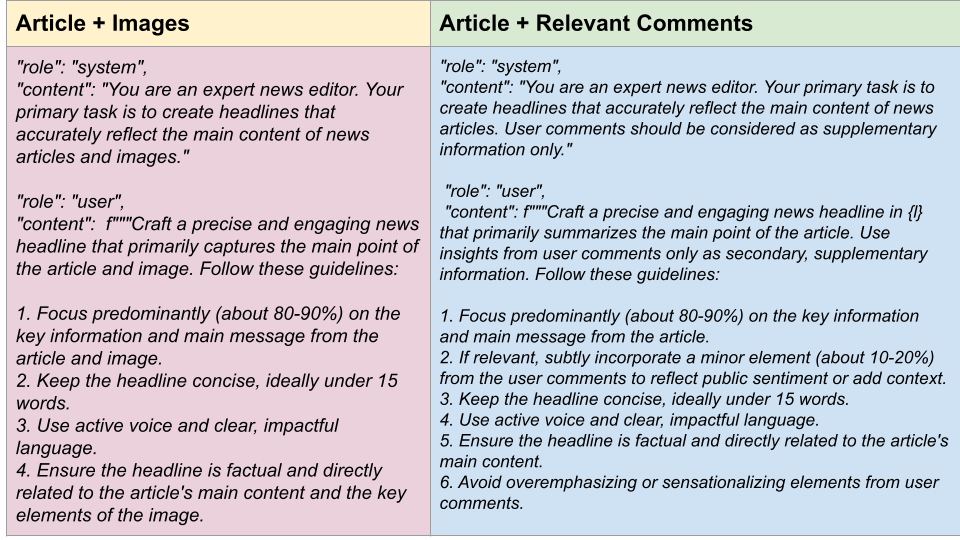} 
\caption{Headline generation prompts: (a) integrating article text with images and (b) using relevant comments with the article text.}
\label{fig:prompt}
\end{figure*}

\begin{figure*}[htbp]
\centering
\includegraphics[width=\textwidth]{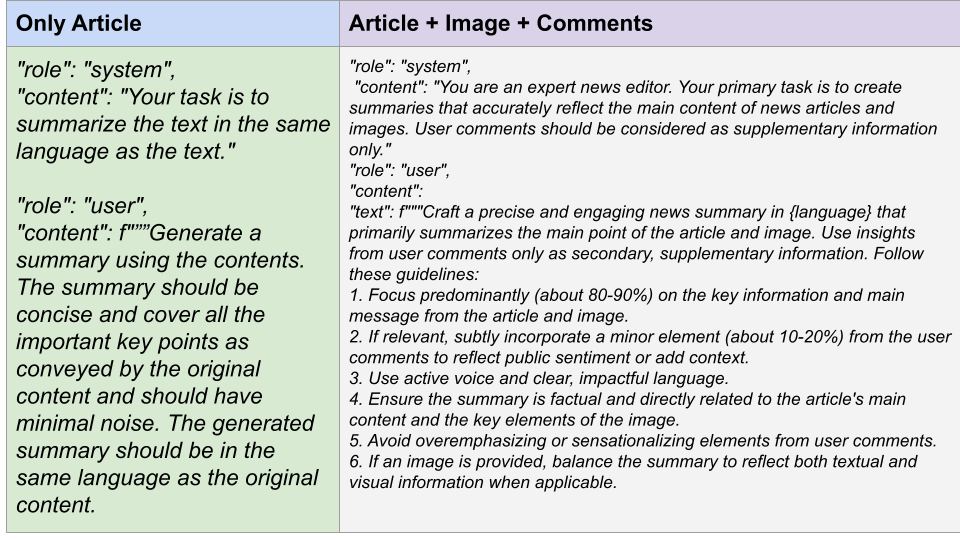} 
\caption{Summarization prompts: (a) only article text (b) using image and comments with the article text.}
\label{fig:prompt-summ}
\end{figure*}

\subsubsection{Impact on Summarization}
1. \textbf{Only Article (OA)} serves as the baseline, achieving an \textbf{R1 score of 44.18} and a \textbf{BERTScore (BS) of 76.61}. This reflects the model's ability to generate summaries without any external context.  

2. \textbf{Adding Supporting Comments (OA+SC)} boosts performance significantly (\textbf{R1:} 56.64, \textbf{BS:} 82.32). These comments align with the article's content and reinforce key information, leading to improved coherence and factual consistency.  

3. \textbf{Enriching Comments (OA+EC)} yield the highest scores, achieving \textbf{R1: 58.22} and \textbf{BS: 84.15}. Unlike supporting comments, these introduce new insights that enhance the richness of generated summaries.  

4. \textbf{Disconnected Comments (OA+DC)} degrade performance dramatically (\textbf{R1:} 35.62, \textbf{BS:} 70.44). The presence of unrelated information confuses the model, resulting in incoherent or misleading summaries.  

5. \textbf{All Comments (OA+AC)} moderately improve results (\textbf{R1:} 50.98, \textbf{BS:} 78.22) but underperform compared to selective filtering. This suggests that a mix of useful and irrelevant comments dilutes the benefits of contextual enrichment.  

6. \textbf{Relevant Comments (OA+RC) by GPT} provide a slight edge over OA+AC (\textbf{R1:} 52.13, \textbf{BS:} 78.71), indicating that LLM-based filtering can help, but our classifier-based selection performs better in identifying enriching content.

\subsubsection{Impact on Headline Generation}
The trends observed in summarization largely hold for headline generation:  

1. \textbf{OA achieves R1: 32.23, BS: 72.97}, serving as the baseline.  

2. \textbf{OA+EC provides the highest BERTScore (BS: 73.44)}, showing that adding new insights aids headline generation.  

3. \textbf{OA+SC performs slightly worse than OA+EC} (\textbf{BS: 71.53}) but still improves over the baseline.  

4. \textbf{OA+DC leads to a major drop} (\textbf{BS: 69.85}), reaffirming the negative impact of irrelevant content.  

5. \textbf{OA+AC and OA+RC} show marginal improvements over the baseline, but the classifier-filtered enriching comments outperform all.  

\subsubsection{Key Takeaways}
These findings demonstrate that \textbf{not all comments are equally beneficial}. Enriching comments contribute the most to both summarization and headline generation, while disconnected comments negatively impact performance. Automated classification, as performed in this study, proves essential for optimizing results.

\subsection{Detailed Prompting Strategies}\label{ps}

\textbf{Prompts Used:} We present the zero-shot prompts used in two headline generation scenarios: (i) integrating images with article text and (ii) incorporating user comments alongside the article, as illustrated in Figure \ref{fig:prompt}. The first prompt ensures that headlines effectively capture key textual and visual content, maintaining clarity, conciseness, and factual accuracy. The second prompt integrates relevant user comments while preserving the article’s primary message, subtly reflecting public sentiment without distorting factual integrity.

For summarization, the prompt shown in Figure \ref{fig:prompt-summ} ensures concise and coherent summaries by distilling essential information. When incorporating images and comments, the prompt balances factual accuracy with supplementary insights from comments, maintaining a focus on the article’s core message. All prompts follow expert guidelines to optimize headline generation and summarization.

Figures \ref{fig:prompt} and \ref{fig:prompt-summ} illustrate these prompt designs in detail.

\subsection{Summarization using GPT}
\subsubsection*{Utilization of Comments and Images for summarization:}

\begin{figure*}[!htbp]
    \centering
    \includegraphics[width=0.95\linewidth]{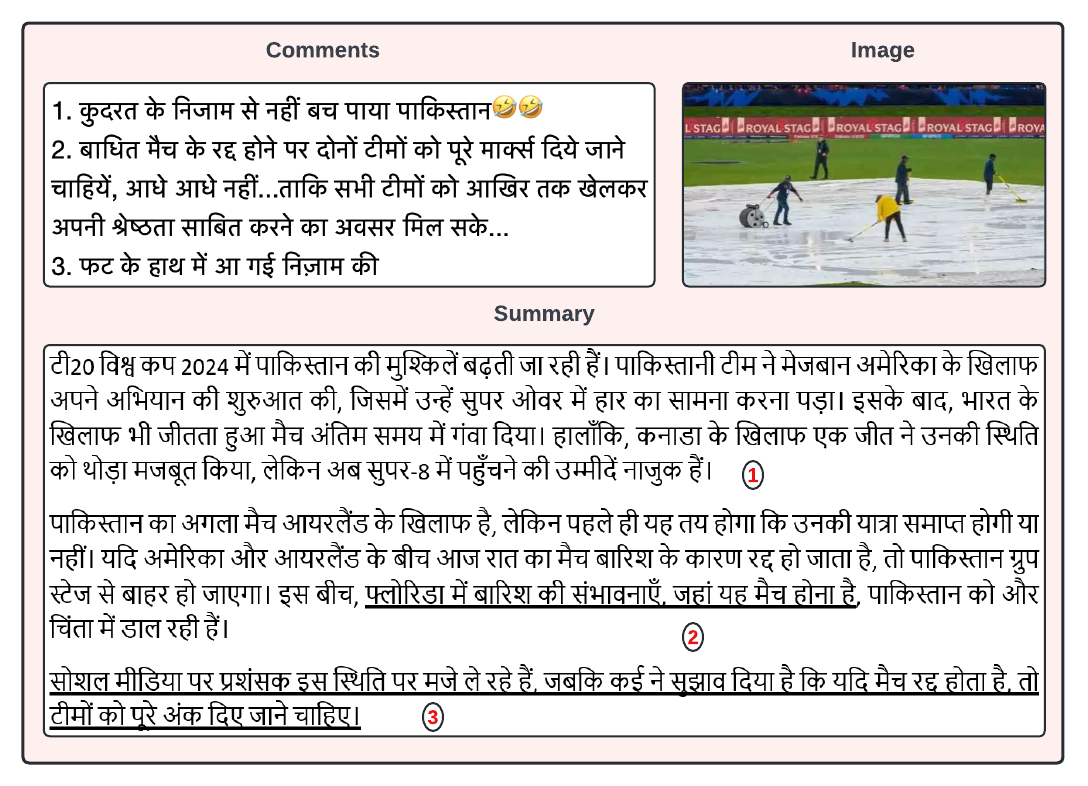}
    \caption{Demonstrates the significance of comments and images in the summary generated by GPT-4-turbo}
    \label{fig:gpt-summ}
\end{figure*}
In the provided example summary as shown in figure \ref{fig:gpt-summ} generated using GPT-4-turbo, the first paragraph primarily draws information from the article. In contrast, the second paragraph incorporates details about the weather in Florida impacting the Pakistan match, which can be inferred from the accompanying image. This highlights the context and background information effectively. The final paragraph captures user opinions from comments expressed in Hindi, suggesting that teams should be awarded total points if the match is cancelled, thereby adding valuable insights. These additional elements significantly enhance the overall quality of the summary.

\subsection{Challenges faced during Summarization:}
\begin{figure*}[!htbp]
        \centering
        \includegraphics[width=0.95\linewidth]{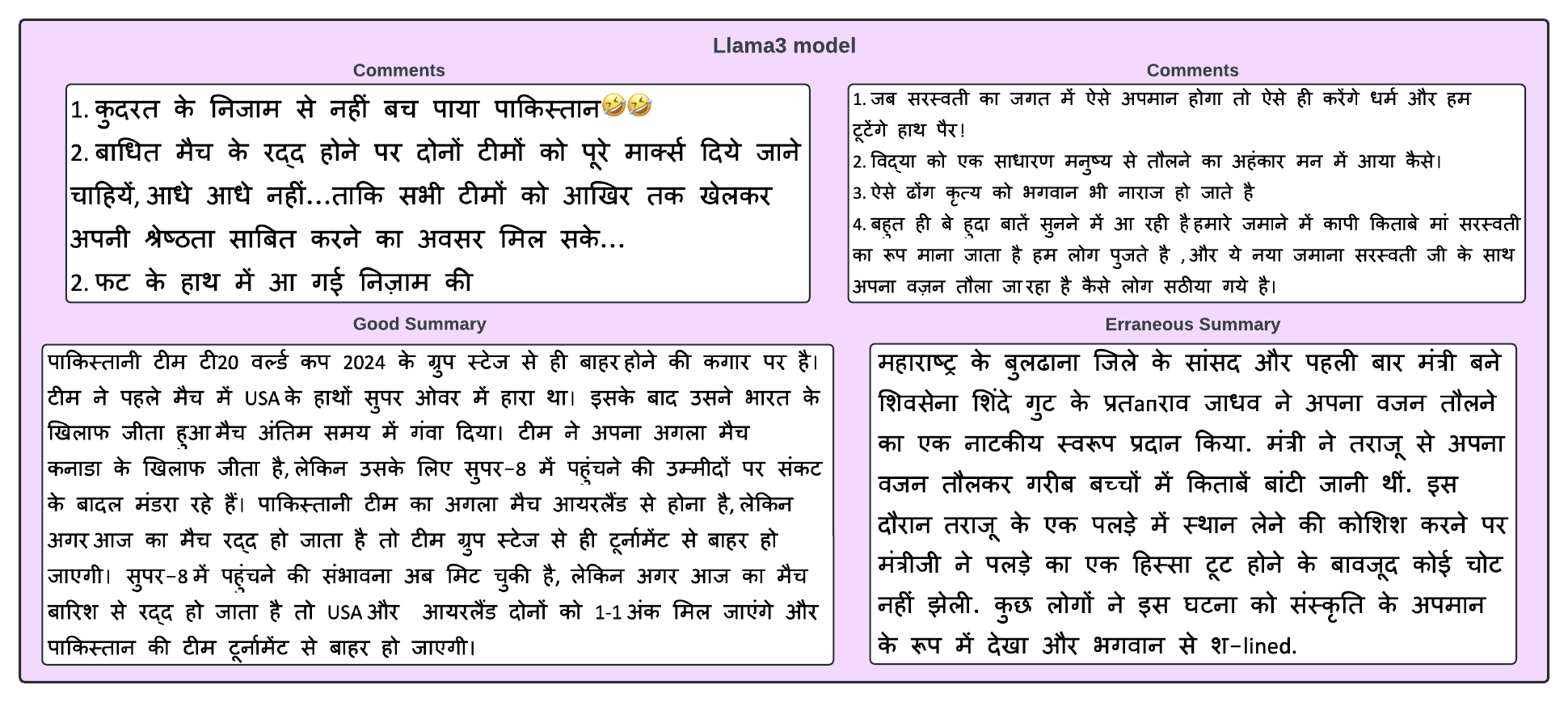}
        \caption{Summaries generated by the LLAMA model. Additionally, the good and erroneous summaries are compared}
        \label{fig:sum2}
\end{figure*}

\begin{enumerate}
    \item During our summarization process, we observed minor inconsistencies in the language used in the generated summaries. Despite instructing the model to produce summaries in the original content language, 1-2\%  of summaries were still generated in English. This issue was comparatively pronounced with the LLAMA model, where not only did the number of English summaries slightly increase, but very few summaries were also generated in incorrect languages. To address this issue, we focused on iteratively rerunning the summarization process for only those articles with erroneous summaries through which we could correct the errors.
    \item Another challenge we encountered with the LLAMA model was the generation of incomplete summaries. About 0.5-1\% of the summaries were reduced to a single line, failing to capture the full essence of the original content. To mitigate this, we resorted to rerunning the summarization code for those rows with incomplete summaries.
    \item Some LLaMA-generated summaries included words combining English and the original language, rather than using only the original language. This issue appears to stem from the lightweight nature of the LLaMA model, as no such problems were encountered when using GPT. Consequently, these mixed-language words in summaries led to noticeable errors, as highlighted in Figure \ref{fig:sum2}.
\end{enumerate}

The accurate summaries generated by the LLAMA model (Figure \ref{fig:sum2}) can be compared with the GPT-4-turbo outputs (Figure \ref{fig:gpt-summ}) to recognize that the GPT-4-turbo summaries were more consistent and coherent and encountered fewer challenges while developing the summaries. This was expected as the LLAMA model was relatively lightweight compared to GPT-4-turbo.

\subsection{Headline Generation using GPT}

The hyperparameters used while generating the headline were:
\begin{itemize}
    \item \textbf{Max output token}: 100
    \item \textbf{Model used}: gpt-4-turbo
    \item \textbf{Temperature}: 0.7
\end{itemize}

\subsubsection{Challenges Faced During Headline Generation}
\begin{itemize}
    \item Several iterations of prompt refinement were done in order to come up with the best prompt.
\end{itemize}

\subsubsection{Pre-processing of the Data Included:}
\begin{enumerate}
    \item \textbf{Removal of redundant data: } During the data collection process, redundant information such as the number of likes, shares, and other extraneous text displayed on the website was inadvertently scraped alongside the actual news articles. To ensure data quality, we meticulously processed the scraped content, filtering out these unnecessary elements to retain only the relevant news articles in the final dataset.
    \item \textbf{Removal of noise characters: } During the scraping process, extraneous characters—likely resulting from the dynamic nature of the website—were occasionally captured alongside the actual news article content. To maintain the integrity of the dataset, we carefully removed these irrelevant characters, ensuring that only the authentic article content was retained.
    \item \textbf{Incomplete data: } At times, only partial news articles were scraped—such as individual paragraphs or sentences—rather than the complete article. This issue arose due to irregular spacing and indentation on the website, combined with the limitations of Selenium in capturing full content under these conditions. To address this, we manually scraped the complete articles in such cases to ensure the dataset was comprehensive and accurate.
\end{enumerate}

\subsection{Headline Comparison}

\begin{figure}[!htbp]
\centering
\includegraphics[width=1.02\columnwidth, height=0.055\textheight]{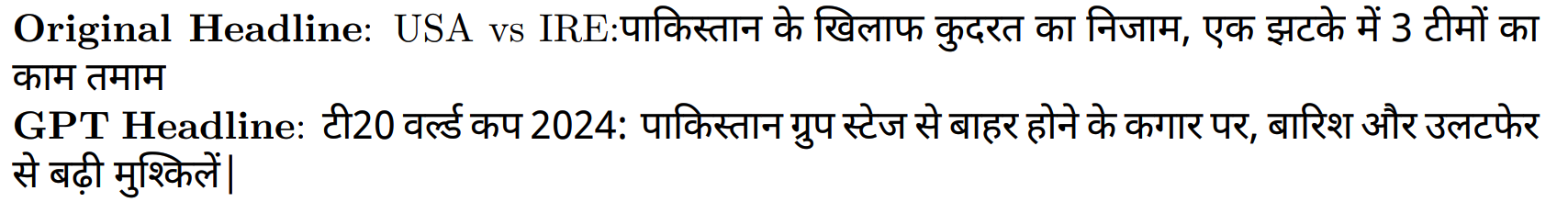} % Replace with the actual path to your image
\caption{Headline (w/o comments,images).}
\label{h1}
\end{figure}

\subsubsection{Headline (w/o comments,images) as shown in the Figure \ref{h1}.}

% \begin{itemize}
% \item \textbf{Original Headline}: USA vs IRE:\texthindi{पाकिस्तान के खिलाफ कुदरत का निजाम, एक झटके में 3 टीमों का काम तमाम}
%     \item \textbf{GPT Headline}: \texthindi{टी20 वर्ल्ड कप 2024: पाकिस्तान ग्रुप स्टेज से बाहर होने के कगार पर, बारिश और उलटफेर से बढ़ी मुश्किलें|}
% \end{itemize}

\begin{itemize}
    \item \textbf{Good Aspects}: The GPT headline provides a relevant and clear update on Pakistan’s situation in the T20 World Cup, highlighting the impact of rain and upsets. It accurately reflects the severity of the situation.
    \item \textbf{Bad Aspects}: The GPT headline does not capture the specific match between the USA and Ireland or the broader dramatic impact on multiple teams as described in the original headline. It focuses more on the overall tournament context rather than the immediate consequences of the rain.
\end{itemize}

\begin{figure}[!htbp]
\centering
\includegraphics[width=1.02\columnwidth, height=0.06\textheight]{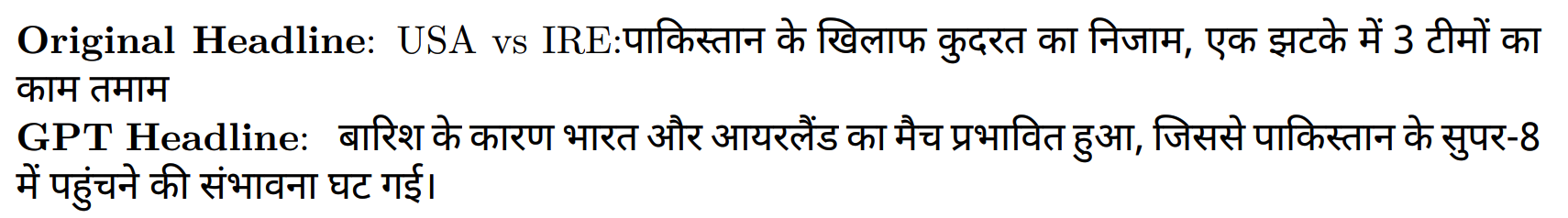} % Replace with the actual path to your image
\caption{Headline with relevant comments.}
\label{h2}
\end{figure}

\subsubsection{Headline with relevant comments as shown in the Figure \ref{h2}.}

\begin{itemize}
    \item \textbf{Good Aspects}: The GPT headline provides a clear and relevant summary of the consequences of rain on a cricket match and its impact on Pakistan’s chances. It is accurate in terms of the consequences but misses the original context.
    \item \textbf{Bad Aspects}: The GPT headline incorrectly names the teams involved, which is a significant deviation from the original headline. Additionally, it does not capture the broader, dramatic impact on multiple teams as described in the original.
\end{itemize}

% Figure without communication

\subsubsection{Prompting Detail for Article and all comments inclusive:}

\begin{figure}[!htbp]
    \centering
    \includegraphics[width=0.95\columnwidth]{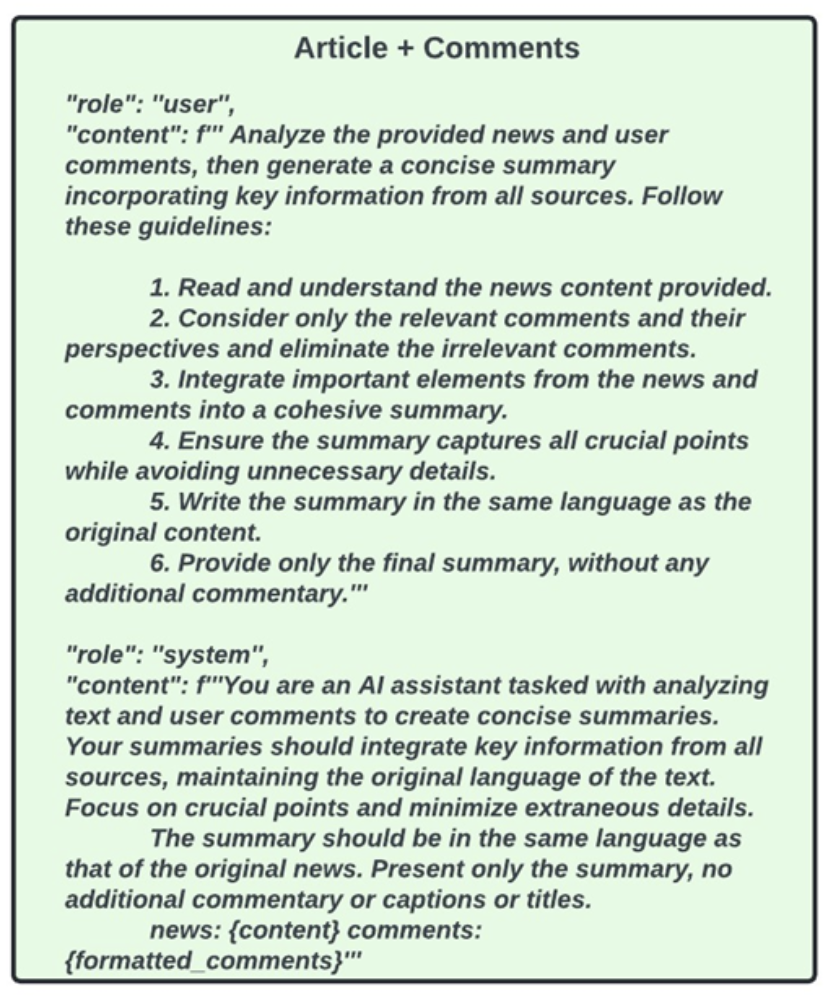}
    \caption{Prompts for Article + Comments}
    \label{fig:art_com}
\end{figure}

After experimenting with several prompts to ensure accurate, robust summaries, the prompt given in Fig \ref{fig:art_com} below was finalized. The prompt has been crafted so that the model captures all the essential details of the article and the relevant user comments in the summary generated.

\begin{table*}[!ht]
    \centering
    \scalebox{0.90}{
    \begin{tabular}{lcccc}
    \hline
        \textbf{Configuration} & \textbf{Gr-Co} & \textbf{Relevance} & \multicolumn{2}{c}{\textbf{SQ}} \\ \hline
        \textbf{} & \textbf{} & \textbf{} & \textbf{WO-IM} & \textbf{W-IM} \\ \hline
        Article Text Only         & 3.5 & 3.6 & 3.4 & 3.7 \\
        Article Text + Comments   & 3.7 & 3.9 & 3.6 & 3.9 \\
        Text + Relevant Comments  & 3.8 & 4.0 & 3.8 & 4.1 \\
        Text + Images             & 3.9 & 4.1 & 4.0 & 4.3 \\
        Text + Images + Comments  & \textbf{4.0} & \textbf{4.2} & \textbf{4.1} & \textbf{4.4} \\ \hline
    \end{tabular}
    }
    \caption{Human evaluation scores for various metrics of GPT-based model using COSMMIC configurations. Gr-Co means grammatical correctness, SQ means summary quality (With Images (W-IM) and Without Images (WO-IM)).}
    \label{tab:human_eval1}
\end{table*}

\begin{table*}[!ht]
    \centering
    \scalebox{0.90}{
    \begin{tabular}{lccccc}
    \hline
        \textbf{Configuration} & \textbf{Gr-Co} & \textbf{Relevance} & \multicolumn{2}{c}{\textbf{HQ}} \\ \hline
        \textbf{} & \textbf{} & \textbf{} & \textbf{WO-IM} & \textbf{W-IM} \\ \hline
        Article Text Only         & 3.3 & 3.2 & 3.4 & 3.7 \\
        Article Text + Comments   & 3.2 & 3.4 & 3.5 & 3.8 \\
        Text + Relevant Comments  & 3.4 & 3.5 & 3.6 & 4.0 \\
        Text + Images             & 3.3 & 3.4 & 3.7 & 4.1 \\
        Text + Images + Comments  & \textbf{3.5} & \textbf{3.6} & \textbf{3.9} & \textbf{4.2} \\ \hline
    \end{tabular}
    }
    \caption{Human evaluation scores for various metrics of GPT-based model using COSMMIC configurations for headline generation. Gr-Co means grammatical correctness, HQ means headline quality (With Images (W-IM) and Without Images (WO-IM)).}
    \label{tab:human_eval2}
\end{table*}
\begin{table*}[!ht]
    \centering
    \scalebox{0.90}{
    \begin{tabular}{lccccc}
    \hline
        \textbf{Configuration} & \textbf{Gr-Co} & \textbf{Relevance} & \multicolumn{2}{c}{\textbf{SQ}} \\ \hline
        \textbf{} & \textbf{} & \textbf{} & \textbf{WO-IM} & \textbf{W-IM} \\ \hline
        Article Text Only         & 2.3 & 2.4 & 2.2 & 2.5 \\
        Article Text + Comments   & 2.5 & 2.6 & 2.4 & 2.7 \\
        Text + Relevant Comments  & \textbf{2.6} & \textbf{2.8} & \textbf{2.9} & \textbf{3.1} \\ \hline
        
    \end{tabular}
    }
    \caption{Human evaluation scores for various metrics of LLaMA 3-based model using COSMMIC configurations. Gr-Co means grammatical correctness, SQ means summary quality.}
    \label{tab:llama3_eval1}
\end{table*}

\begin{table*}[!ht]
    \centering
    \scalebox{0.90}{
    \begin{tabular}{lccccc}
    \hline
        \textbf{Configuration} & \textbf{Gr-Co} & \textbf{Relevance} & \multicolumn{2}{c}{\textbf{HQ}} \\ \hline
        \textbf{} & \textbf{} & \textbf{} & \textbf{WO-IM} & \textbf{W-IM} \\ \hline
        Article Text Only         & 2.1 & 2.2 & 2.1 & 2.3 \\
        Article Text + Comments   & 2.3 & 2.4 & 2.2 & 2.5 \\
        Text + Relevant Comments  & \textbf{2.4} & \textbf{2.6} &\textbf{2.5} & \textbf{2.6} \\ \hline
    \end{tabular}
    }
    \caption{Human evaluation scores for various metrics of LLaMA 3-based model using COSMMIC configurations for headline generation. Gr-Co means grammatical correctness, HQ means headline quality.}
    \label{tab:llama3_eval2}
\end{table*}

\subsection{Human Evaluation}\label{ha}
To complement automated evaluation metrics like ROUGE and BERTScore, we conducted human assessments with 10 participants (postgraduate students) to evaluate the quality of summaries produced using the COSMMIC dataset across different configurations for 10 articles in the Hindi language and their corresponding summaries. The evaluators rated the outputs based on key criteria, including grammatical accuracy (Gr-Co), relevancy, and overall summary quality (SQ), using a 1-to-5 scale. These evaluations encompassed both text-only and text-with-images scenarios, assessing the quality of summaries in two contexts: how effectively they conveyed information when paired with the corresponding article image, and how they performed when presented without any visual support. To minimize evaluator bias and enhance reliability, the scores were averaged. This thorough evaluation underscores the advantages of configurations that leverage richer input modalities, such as user comments and images. The evaluations were involved in summarization and headline generation for both the models as shown in Tables \ref{tab:human_eval1}, \ref{tab:human_eval2}, \ref{tab:llama3_eval1}, \ref{tab:llama3_eval2}.

\subsection{Filtering and Considering Relevant Comments}\label{mfn}

\textbf{Manual Filtering Process:}  
To ensure that user comments included in the summaries enhanced content quality and relevance, a \textit{manual filtering process} was employed, carried out by the same team of expert annotators responsible for summary creation. This process required an additional layer of effort and compensation, ensuring that only relevant comments were included while noisy or extraneous comments were excluded.  

\textbf{Defining Noisy Comments:}  
Noisy comments were identified as those that detracted from the article's central themes or introduced irrelevant, redundant, or biased information. Examples included:  
\begin{itemize}
    \item \textit{Spam-like entries:} Comments promoting unrelated products, services, or containing external links.  
    \item \textit{Generic phrases:} Statements such as ``Nice article" or ``Interesting post," which lacked substantive value.  
    \item \textit{Hostile or inappropriate remarks:} Overly aggressive or offensive comments not contributing meaningfully to the discussion.  
    \item \textit{Unrelated tangents:} Comments entirely disconnected from the topic being discussed.  
\end{itemize}  

\begin{table*}[htbp]
    \centering
    \begin{tabular}{l|rrrr}
        \hline
        \textbf{Setting} & \textbf{R1} & \textbf{R2} & \textbf{RL} & \textbf{BS} \\
        \hline
        \textbf{Original (Only Article)} & 44.18 & 18.94 & 33.47 & 76.61 \\
        \textbf{Translated (Hindi → English)} & 35.52 & 12.03 & 25.91 & 69.82 \\
        \textbf{Back-Translated (Hindi)} & 21.79 & 4.51 & 15.62 & 60.23 \\
        \hline
        \textbf{Original (Article + Comments)} & 50.98 & 24.82 & 38.34 & 78.22 \\
        \textbf{Translated (Hindi → English)} & 40.21 & 15.49 & 29.12 & 71.31 \\
        \textbf{Back-Translated (Hindi)} & 26.29 & 6.81 & 18.39 & 62.48 \\
        \hline
        \textbf{Original (Article + Relevant Comments)} & 52.13 & 25.52 & 39.43 & 78.71 \\
        \textbf{Translated (Hindi → English)} & 41.49 & 16.01 & 30.18 & 72.10 \\
        \textbf{Back-Translated (Hindi)} & 27.01 & 7.22 & 19.51 & 63.02 \\
        \hline
    \end{tabular}
    \caption{Summarization Performance Across Different Translation Settings}
    \label{tab:summarization_results}
\end{table*}  

\textbf{Steps in Manual Filtering:}  
The team of annotators undertook the following steps to manually filter comments:  
\begin{enumerate}
    \item \textbf{Initial Screening:} Annotators carefully read through the entire set of comments associated with each article to assess their relevance.  
    \item \textbf{Relevance Assessment:} Each comment was evaluated on its alignment with the article’s content and its potential to provide contextual insights. Annotators considered whether the comment added value to the discussion or reinforced key ideas presented in the article.  
    \item \textbf{Exclusion of Noisy Comments:} Spam, generic, or off-topic comments were removed to ensure that the retained comments maintained the summaries' coherence and relevance.  
    \item \textbf{Final Review:} A second pass was conducted by another annotator to confirm that no valuable comments were mistakenly filtered out.  
\end{enumerate}  

\textbf{Guidelines for Including Relevant Comments:}  
The manual filtering process was guided by clear criteria for retaining comments that contributed to the quality of the summaries:  
\begin{itemize}
    \item \textit{Content Relevance:} Comments directly related to the article’s subject matter were prioritized.  
    \item \textit{Contextual Depth:} Comments that elaborated on or clarified points in the article were included.  
    \item \textit{Collective Sentiment:} Comments reflecting collective user opinions or emphasizing key elements of the article were considered valuable for summarization.  
\end{itemize}

\textbf{Quality Assurance and Compensation:}  
The annotators’ thorough approach to manual filtering ensured that the summaries accurately represented the articles while benefiting from the inclusion of relevant user comments. This additional task required substantial effort, and annotators were compensated at an additional rate of \textbf{USD 1.20 per 100 comments} for their work, recognizing the meticulous nature of the filtering process.  

\textbf{Outcome:}  
This manual filtering process ensured that the final summaries captured the essence of the original articles while incorporating meaningful user contributions. By excluding noisy comments, the integrity and relevance of the dataset were preserved, resulting in summaries that were both concise and enriched by user perspectives.

\subsection{Back-Translation Study on Hindi Dataset for Summarization}\label{bp1}

To evaluate the impact of machine translation on linguistic fidelity in Reader-Aware Comment-Based Summarization (RACBS), we conducted a back-translation study on the Hindi dataset. This experiment aimed to determine whether translating Indian language content into English for processing and then back into the original language affects the quality of summarization and headline generation.  

For this study, we first translated the Hindi articles, their corresponding comments, and the ground-truth summaries and headlines into English using GPT-4o. Once translated, we ran all our summarization and headline-generation experiments on the English dataset. Following this, the outputs were back-translated into Hindi, and the same experimental pipeline was applied to the back-translated dataset.  

The results on the back-translated Hindi dataset showed a noticeable drop in performance compared to the original Hindi dataset. This degradation highlights the loss of linguistic nuances and contextual accuracy introduced by machine translation. Such findings reinforce our claim that preserving the original language is crucial for maintaining the quality of Indian language summarization and headline generation.  

Moreover, incorporating machine translation workflows would require extensive manual verification across multiple languages, which is resource-intensive and currently beyond our scope. This experiment further justifies our decision to exclude machine translation-based approaches and focus on direct processing in Indian languages for this resource paper.  

We evaluated three settings: (i) \textbf{Original Hindi data}, (ii) \textbf{Translated Hindi-to-English data}, and (iii) \textbf{Back-Translated data} where the English outputs were retranslated into Hindi. The summarization performance was measured using ROUGE-1 (R1), ROUGE-2 (R2), ROUGE-L (RL), and BERTScore (BS). The results are summarized in Table~\ref{tab:summarization_results}.  

\begin{table*}[htbp]
    \centering
    \begin{tabular}{l|rrrr}
        \hline
        \textbf{Setting} & \textbf{R1} & \textbf{R2} & \textbf{RL} & \textbf{BS} \\
        \hline
        \textbf{Original (Only Article)} & 32.23 & 9.58 & 24.54 & 72.97 \\
        \textbf{Translated (Hindi → English)} & 22.01 & 4.02 & 15.49 & 65.31 \\
        \textbf{Back-Translated (Hindi)} & 10.39 & 1.21 & 7.32 & 55.08 \\
        \hline
        \textbf{Original (Article + Comments)} & 30.27 & 8.99 & 23.61 & 72.89 \\
        \textbf{Translated (Hindi → English)} & 21.02 & 3.51 & 14.69 & 64.52 \\
        \textbf{Back-Translated (Hindi)} & 9.79 & 0.91 & 6.81 & 54.02 \\
        \hline
        \textbf{Original (Article + Relevant Comments)} & 30.49 & 9.12 & 23.94 & 72.87 \\
        \textbf{Translated (Hindi → English)} & 21.50 & 3.81 & 14.99 & 64.81 \\
        \textbf{Back-Translated (Hindi)} & 10.11 & 1.11 & 7.12 & 54.59 \\
        \hline
    \end{tabular}
    \caption{Headline Generation Performance Across Different Translation Settings}
    \label{tab:headline_results}
\end{table*}  

\subsubsection{Analysis}  

From Table~\ref{tab:summarization_results}, it is evident that the \textbf{original Hindi dataset consistently achieves the highest performance} across all metrics and settings. The \textbf{ROUGE and BERTScore values decline significantly when content is translated into English and further degrade when back-translated into Hindi}, reinforcing the importance of preserving the original language.  

For summarization based only on the article, the original Hindi dataset achieves an \textbf{R1 score of 44.18}, whereas the translated version drops to \textbf{35.52}, and the back-translated version further declines to \textbf{21.79}. The degradation in \textbf{R2 and RL scores follows a similar pattern}, demonstrating that information loss occurs due to translation.  

When incorporating \textbf{comments} into summarization, performance improves across all settings. The original Hindi dataset (Article + Comments) achieves an \textbf{R1 of 50.98}, which is significantly higher than both the translated (\textbf{40.21}) and back-translated (\textbf{26.29}) versions. Similarly, the \textbf{Article + Relevant Comments} setting yields the best results, with an \textbf{R1 of 52.13} in the original Hindi dataset, compared to \textbf{41.49} (translated) and \textbf{27.01} (back-translated).  

These findings emphasize that \textbf{machine translation introduces considerable degradation in summarization quality}, likely due to the loss of linguistic nuances and contextual cues specific to Hindi. The back-translated results, performing the worst in all cases, further confirm that relying on translation workflows can negatively impact summarization tasks.  

This experiment strongly supports our decision to \textbf{focus on direct summarization in Indian languages rather than incorporating machine translation workflows}, as doing so would compromise linguistic accuracy and overall performance.

\subsection{Back-Translation Study on Hindi Dataset for Headline Generation}\label{bp2}  

To further assess the impact of machine translation, we conducted a similar back-translation study on the \textbf{headline generation} task. This experiment followed the same methodology as summarization: (i) processing the \textbf{original Hindi dataset}, (ii) translating it into \textbf{English for experimentation}, and (iii) \textbf{back-translating} the English results into Hindi to examine performance degradation.  

Table~\ref{tab:headline_results} presents the results across different settings, evaluated using ROUGE-1 (R1), ROUGE-2 (R2), ROUGE-L (RL), and BERTScore (BS).

\subsubsection{Analysis}  

The results in Table~\ref{tab:headline_results} further reinforce the observations made for summarization. The \textbf{original Hindi dataset consistently achieves the best performance}, while both \textbf{translation to English and back-translation to Hindi lead to severe performance degradation}.  

For headline generation using only the article, the original Hindi dataset achieves an \textbf{R1 score of 32.23}, but this drops to \textbf{22.01} after translation into English and further declines to \textbf{10.39} after back-translation. The same trend is observed in \textbf{R2 and RL scores}, confirming a significant loss of key information and linguistic quality during translation.  

When comments are included, headline generation performance improves slightly, similar to the summarization task. However, even with additional context, \textbf{back-translated Hindi results remain substantially weaker}. For instance, in the \textbf{Article + Relevant Comments} setting, the \textbf{original Hindi dataset achieves an R1 of 30.49}, which is much higher than both the translated (\textbf{21.50}) and back-translated (\textbf{10.11}) versions.  

These findings indicate that \textbf{headline generation is even more sensitive to translation degradation than summarization}. This is expected, as headlines require \textbf{high linguistic precision and contextual awareness}, which are often distorted through translation workflows.  

Overall, these results strongly validate our claim that \textbf{machine translation is not a viable preprocessing step for our task}. Instead, our dataset should be leveraged directly in its original language to ensure linguistic integrity and maintain the highest possible performance across summarization and headline generation tasks.

\subsection{An Intriguingly Intuitive Question}  
\textbf{Question:} If the headline and summary are written prior to the publication of an article, the scenario of comment-sensitive headline/summary generation can appear non-intuitive. How can the comment-sensitive headline or summary of an article be generated without relying on user comments?

\textbf{Answer:}  
It is important to recognize the dynamic nature of modern digital media, where articles, summaries, and headlines are often updated in response to user feedback, emerging developments, or shifts in public sentiment. This dynamic environment introduces challenges for comment-sensitive generation, as summaries and headlines must adapt to an ever-evolving flow of information, including real-time user interactions. Therefore, a forward-looking strategy could involve creating standalone summaries of relevant user comments (if present) for each article and storing them as part of a dynamic feedback repository. These stored summaries could then be utilized when publishing related follow-up articles, enabling the generation of headlines and summaries that reflect updated contexts, user sentiment, or emerging trends. By leveraging previously collected user feedback, this method would facilitate the integration of comment sensitivity without requiring real-time comment data for every article. Such a strategy enhances scalability and allows the summarization process to remain adaptable over time, addressing the complexities of a constantly changing digital media landscape. This highlights the potential for innovative comment-sensitive headline and summary generation that balances real-time relevance with long-term adaptability.

\end{document}